# Growth-Inspired Graph Generation and Inverse Design of Mechanical Lattices via Dot Matrices Database Augmentation and GCNN

Weiyun Xu,[1*] Jiamu Liu[2]

[1] Department of Civil and Environmental Engineering, University of Illinois Urbana-Champaign, Urbana, Illinois 61801, USA

[2] State Key Laboratory of Clean and Efficient Turbomachinery Power Equipment, Department of Mechanical Engineering, Tsinghua University, Beijing, 100084, China

*Corresponding author:* weiyunxu@illinois.edu

## Abstract

*Natural load-bearing and transport networks are not assembled in a single step; they emerge through a temporally ordered process of growth, branching, reinforcement, and loop formation. Inspired by this developmental logic, this work introduces a morphogenetic graph-generation framework for mechanical lattices in which a discrete dot matrix provides potential nodes and the final architecture is created by sequential cross-layer and intra-layer growth. The same rule is visualized in two dimensions as a leaf-vein-like developmental sequence and implemented in three dimensions on a 3 × 3 × 3 nodal matrix containing 27 candidate nodes. A dataset of distinct three-dimensional lattices was evaluated by beam-based finite element analysis and represented directly as graphs. A graph convolutional neural network (GCNN) with three graph-convolution layers and dual global pooling learns the topology-property mapping and predicts effective compressive stiffness. Coupling the GCNN surrogate with rapid structural sampling enables inverse design: for a target stiffness of 1000 MPa, the selected design was predicted at 1042.43 MPa and validated by finite element analysis at 1027.49 MPa. Beyond straight members, the framework has also been extended to parameterized horseshoe-shaped curved beams made of nonlinear materials, enabling topology-geometry design toward prescribed deformation shapes. Our work provides a paradigm for augmenting the database of mechanical metamaterials, and the resulting perspective links biological morphogenesis, graph learning, and nonlinear shape programming in a unified generative design framework for architected materials.*

## 1. Introduction

Mechanical metamaterials derive their unusual macroscopic response from architecture rather than chemical composition alone. By programming connectivity, slender-member geometry, hierarchy, and internal mechanisms, architected materials can achieve combinations of stiffness, strength, compressibility, multistability, and shape change that are difficult to realize in conventional solids [1-14]. This geometric freedom is also the central difficulty of their design: even a modest lattice domain admits an enormous number of topologies, while nonlinear geometry and material behavior further enlarge the design space.

Data-driven design provides a route through this combinatorial complexity. Machine-learning surrogates can replace repeated high-cost simulations, while inverse-design algorithms can search structural spaces using desired properties rather than manually chosen geometries [15-23]. Recent graph-based approaches are particularly attractive for truss and lattice metamaterials because the native object is already a graph: joints are vertices, members are edges, and both topology and geometry can be represented without rasterizing the structure [24,25]. However, most graph databases still treat each architecture as a finished object sampled from a predefined combinatorial space. The process by which a network is generated is usually secondary.

Nature suggests a different viewpoint. Leaf venation, vascular networks, and other biological distribution systems develop sequentially. Primary pathways emerge first; secondary branches appear as the domain grows; lateral connections and loops then provide redundancy and alternative transport routes. Biologically motivated models of leaf venation explicitly exploit this time ordering, while optimization studies have shown how looped networks can emerge from competing requirements such as efficiency, fluctuating loads, damage tolerance, and growth [26-28]. For mechanical lattices, the developmental process itself can therefore be used as an inductive bias: rather than drawing a complete graph at once, we let the graph grow.

Here we reinterpret the dot-matrix representation in this morphogenetic language and develop it into a dimension-independent growth generator. A two-dimensional leaf-like construction is first used to expose the temporal logic, after which the same operations are applied to a three-dimensional 3 × 3 × 3 nodal scaffold. The resulting lattices are encoded as graphs and evaluated by finite element analysis (FEA). A graph convolutional neural network (GCNN) then learns the mapping from structure to effective stiffness, enabling rapid screening and inverse selection. Importantly, the biological analogy is used at the level of developmental logic - sequential activation, branching, and loop-forming reinforcement - rather than as a literal simulation of plant physiology.

A second objective is to show how this representation can grow beyond straight-member trusses. Flexible and shape-morphing metamaterials increasingly rely on curved beams, large rotations, and nonlinear constituent behavior [29-35]. We therefore introduce a current extension in which graph edges are replaced by parameterized horseshoe-shaped centerlines and coupled to nonlinear mechanics. This expands the design variables from connectivity alone to topology, member curvature, and material response, making target-deformation inverse design possible. Together, these elements establish a unified workflow from morphogenetic generation to learned property prediction and, ultimately, programmable nonlinear shape change.

## 2. Materials and methods

### 2.1. Morphogenetic representation on a discrete dot matrix

Each growth step contains two operations. First, cross-layer growth connects newly activated nodes to nodes in the immediately preceding layer. These parent-child links establish the primary skeleton and create a hierarchical path from the seed toward the expanding front. Second, intra-layer growth introduces lateral links among nodes activated at similar developmental times. These later links play a role analogous to anastomoses in venation: they increase local redundancy, create alternate load paths, and thicken the graph without abandoning the growth history. In the present generator, candidate intra-layer connections are restricted to node pairs separated by less than $\sqrt{3}$ grid units. The distance restriction suppresses nonlocal members and preserves a compact, mechanically meaningful neighborhood.

The biological analogy is intentionally abstract. Classical leaf-venation models can include auxin transport, source depletion, Voronoi partitions, or explicit growth of the leaf blade [26]. Our generator instead extracts only three generic features that are advantageous for architected-material design: temporal ordering, hierarchical branching, and delayed loop formation. This reduction keeps the design variables transparent and produces graphs that can be directly simulated and learned. The generated lattice is schematically shown in Figure 1.

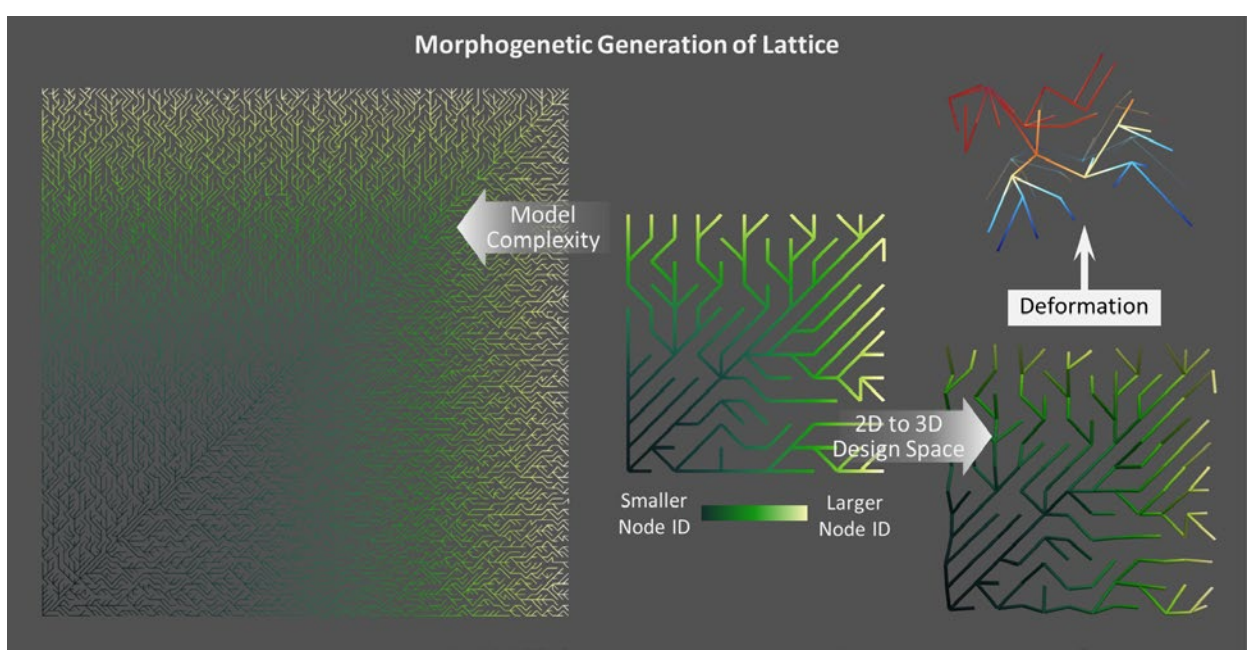

Figure 1: Lattice generation.

The geometric representation of mechanical metamaterials plays a critical role in determining both their achievable property space and the efficiency of subsequent computational modeling. Inspired by hierarchical and adaptive transport networks observed in natural leaf venation systems, we propose a growth-based structural representation method built upon discrete dot matrices. Leaf venation networks exhibit highly efficient load distribution and robustness due to their multi-scale branching patterns, making them a suitable biological analogue for the design of lattice-based mechanical metamaterials.

In this work, each candidate structure is initialized from a three-dimensional nodal matrix of size 3×3×3, yielding a total of 27 discrete nodes. These nodes form the fundamental structural skeleton and are assigned Cartesian coordinates $(x, y, z)$, where each coordinate takes integer values from 0 to 3. This discrete spatial embedding ensures a consistent node count across all generated structures, while allowing for rich topological diversity through variations in connectivity.

The structure generation process consists of two sequential stages: cross-layer generation and intra-layer generation. To facilitate a controlled growth mechanism, nodes are first grouped into concentric layers according to their Euclidean distance from the origin (0, 0, 0). Specifically, nodes sharing similar distances are classified into the same layer, resulting in a layered organization that mimics the hierarchical growth observed in natural vascular systems.

During the cross-layer generation stage, the growth process initiates at the second layer rather than the innermost core. This design choice prevents excessive local densification near the origin and promotes a more uniform distribution of load paths throughout the structure. In this stage, nodes in the current layer are probabilistically connected to nodes in the immediately preceding layer (Fig.

2a–b). These inter-layer connections establish the primary load-transmission pathways and ensure global structural integrity.

Following cross-layer growth, intra-layer connections are introduced to enhance local stiffness and structural redundancy. In this step, nodes within the same layer are allowed to form connections with one another. To maintain mechanical feasibility and avoid unrealistically long or intersecting members, a geometric constraint is imposed: connections are only permitted between node pairs whose Euclidean distance is smaller than $\sqrt{3}$, which corresponds to the maximum diagonal length between adjacent nodes in the discrete grid. This constraint ensures that all structural members are physically realizable and mechanically meaningful.

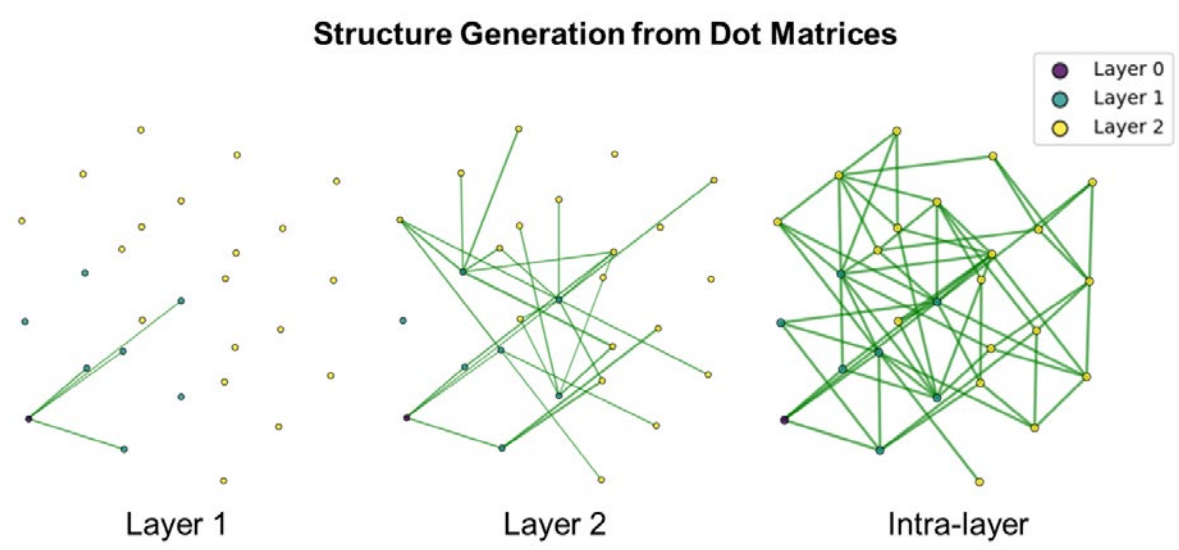

Figure 2: Structure generation from dot matrices.

## 2.2. Graph data and structure of GCNN

Traditional Convolutional Neural Networks (CNNs) have demonstrated remarkable success in image-based tasks due to their ability to exploit spatial locality and translational invariance. However, they are not well suited for modeling lattice or truss-based mechanical metamaterials, whose defining characteristics lie in their connectivity rather than pixel-wise spatial arrangement. In contrast, graph-based representations provide a natural and mathematically consistent framework for encoding such structures.

In this study, each lattice structure generated from the dot-matrix representation is converted into a graph $G(V, E)$, where the set of vertices $V$ corresponds to the structural nodes and the set of edges $E$ represents the beam elements connecting node pairs. Each node is described by a three-dimensional coordinate vector, resulting in a node feature matrix of size $27 \times 3$. Connectivity information is stored separately as an edge list, where each edge is defined by the indices of its two endpoint nodes after sequential numbering.

This graph-based representation allows the use of Graph Convolutional Neural Networks (GCNNs), which generalize the concept of convolution to irregular graph domains. A GCNN model was constructed to learn the nonlinear mapping between lattice topology and its effective mechanical stiffness. The architecture consists of two major components: a graph encoder for feature extraction and a regression head for property prediction.

The graph encoder comprises three graph convolution layers with hidden dimensions of 16, 64, and 128, respectively. These layers progressively transform low-dimensional geometric inputs into higher-level structural descriptors by aggregating information from neighboring nodes. Through this message-passing mechanism, the GCNN captures both local connectivity patterns and global load-transfer pathways. A batch normalization layer is applied after the convolutional layers to stabilize the training process and improve convergence speed. Nonlinearity is introduced via the Rectified Linear Unit (ReLU) activation function, enabling the model to approximate complex, non-linear relationships between structure and property.

Although all structures share an identical number of nodes, their edge connectivity—and thus topology—varies significantly. As a result, the node-level features produced by the graph encoder must be aggregated into a fixed-size global representation before regression. To achieve this, a dual pooling strategy is employed, combining global average pooling and global maximum pooling. While average pooling captures the overall structural trend, max pooling preserves extreme or dominant features that may govern stiffness, such as highly stressed load-bearing paths.

The pooled global feature vector is subsequently passed through a cascade of three fully connected layers with 128, 64, and 32 neurons, respectively. These layers perform nonlinear regression and progressively reduce feature dimensionality. The final output layer consists of a single neuron with linear activation, producing a scalar prediction of structural stiffness. Model training is performed by minimizing the mean absolute error (MAE) loss function using gradient-based optimization.

## 2.3. Finite element methods

To generate accurate mechanical labels for training and validation, finite element analysis (FEA) was conducted using the commercial software ABAQUS (SIMULIA, 2019). Each lattice structure was imported into the simulation environment and discretized using B31 beam elements, which are well suited for modeling slender truss members under combined bending and axial loading. To ensure numerical accuracy, each beam was subdivided into 20 elements along its length.

A uniaxial compression test was simulated to characterize the effective stiffness of each structure. Boundary conditions were applied such that all translational and rotational degrees of freedom of nodes located on the bottom surface were fully constrained. A prescribed displacement of −0.1 mm was applied uniformly to all nodes on the top surface along the z-axis, inducing compressive deformation.

During the simulation, reaction forces and displacements were recorded to construct stress–strain curves for each structure. The effective stiffness was extracted by performing a linear fit over the elastic region of the stress–strain response and identifying the maximum slope (Fig. 3). This stiffness value serves as the ground-truth label for subsequent machine learning tasks.

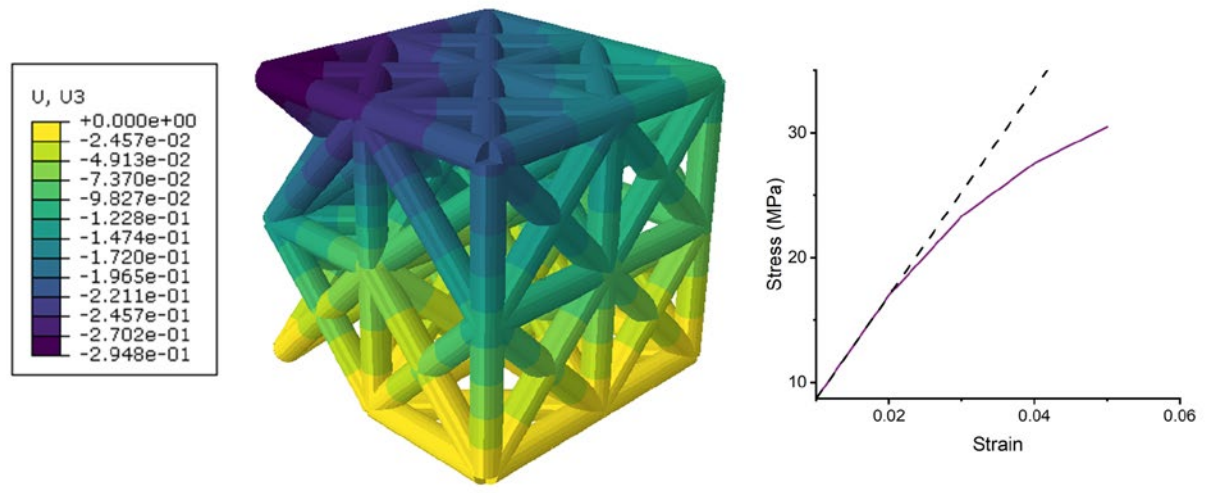

Figure 3: Finite element analysis and stiffness extraction

# 3. Results

## 3.1. Potential property space of current representation

To evaluate the expressive power of the proposed dot-matrix representation, an initial dataset of 1,000 structures was generated and analyzed using FEM. The resulting stiffness distribution is shown in Fig. 4. The data approximately follow a normal distribution, with a mean stiffness of 1098 MPa and a standard deviation of 126.63 MPa.

Although the majority of samples fall within a relatively narrow band, the overall stiffness range spans nearly 20% of the mean value. Notably, the presence of extreme outliers—such as structures exhibiting stiffness as low as 655 MPa—demonstrates that the representation is capable of producing non-trivial and highly distinct mechanical responses. This diversity is essential for both surrogate model training and inverse design, as it ensures coverage of both typical and atypical regions of the property space.

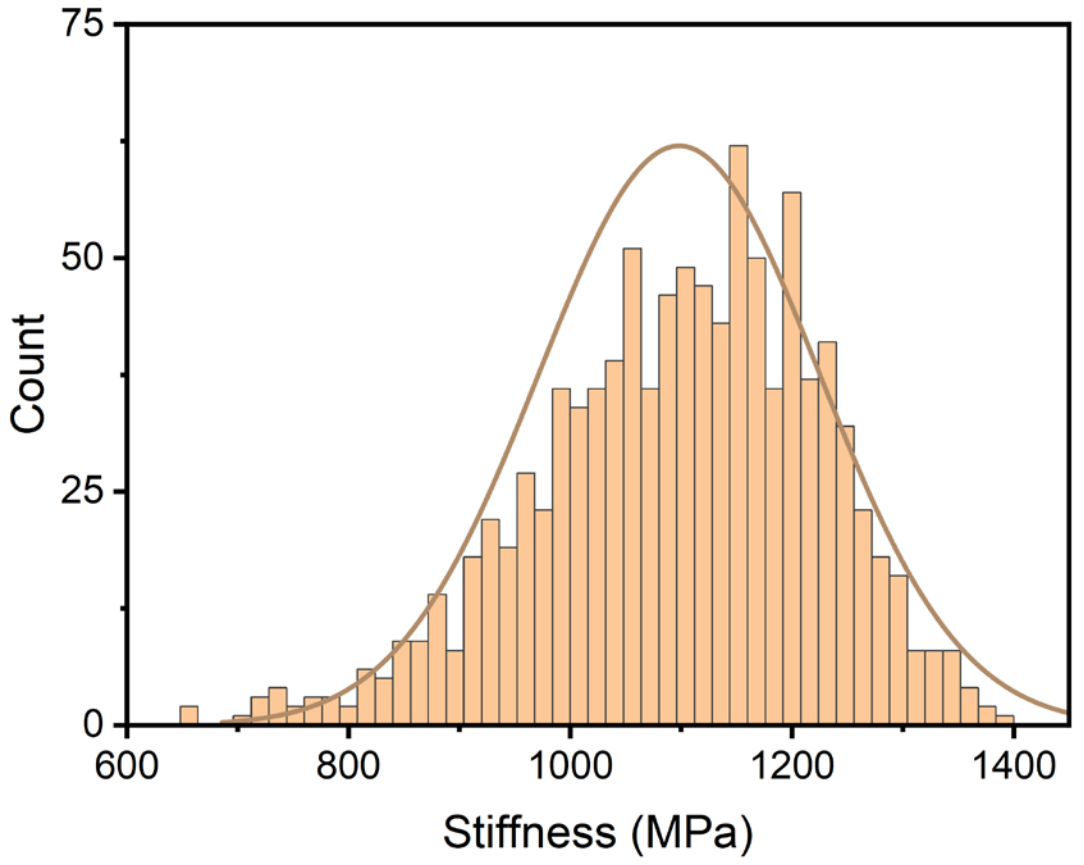

Figure 4. Distribution of the stiffness properties of 1000 samples in initial dataset.

## 3.2. Prediction through GCNN model

The GCNN model was trained on the 1,000-sample dataset using a learning rate of 0.001. Training terminated at the 403rd epoch. The evolution of training and validation losses is shown in Fig. 5a. Both losses decrease steadily during early epochs, indicating effective feature learning. However, after approximately 100 epochs, the validation loss plateaus while the training loss continues to decrease, suggesting the onset of overfitting.

Despite this limitation, the model achieves strong predictive performance. On the validation set, the GCNN reaches a coefficient of determination $R^2 = 0.84$ and an MAE of 37 MPa (Fig. 5b). Most predictions lie within a 5% relative error margin. The model performs particularly well for high-stiffness structures, whereas prediction accuracy deteriorates for low-stiffness samples. This asymmetry is attributed to data imbalance: structures with extremely low stiffness are underrepresented in the training dataset, limiting the model's ability to learn their distinctive features.

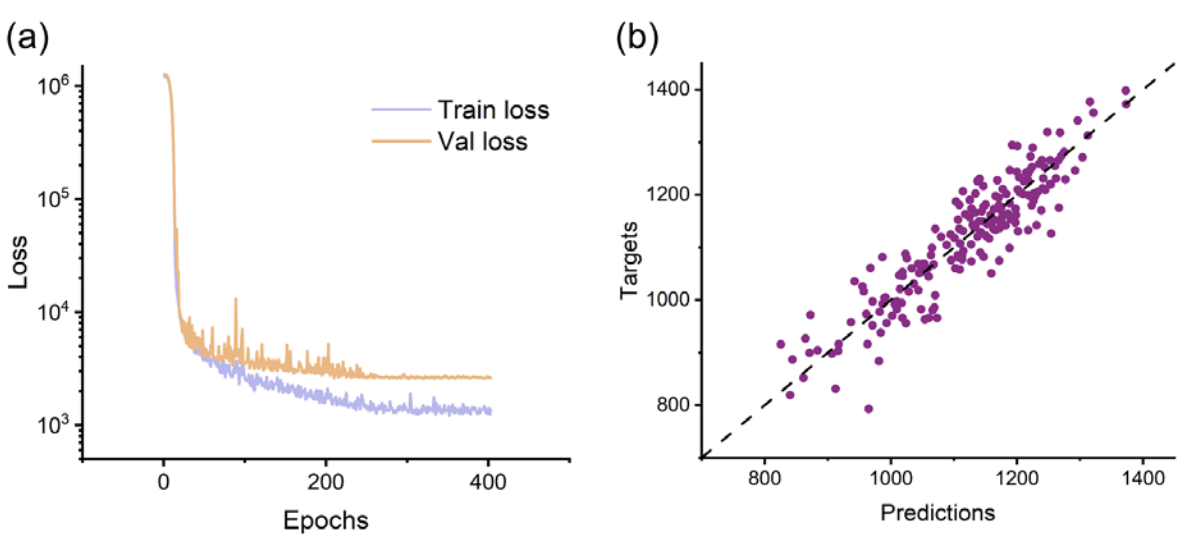

Figure 5. Train losses and validation losses in the model training process (a), and properties predicted by the model vs. ground truths on validation dataset (b).

## 3.3. Inverse design for a prescribed stiffness

Building upon the trained GCNN surrogate model, a simple inverse design framework was implemented to generate structures with target stiffness values. The framework integrates brute-force structural sampling with rapid property prediction. For a specified target stiffness, a large number of candidate structures are generated and evaluated by the GCNN. Candidates whose predicted stiffness falls within ±5% of the target are selected for FEM validation.

As demonstrated in Fig. 6, the framework successfully identified a structure targeting a stiffness of 1000 MPa. The GCNN predicted a value of 1042.43 MPa, while FEM simulation yielded a true stiffness of 1027.49 MPa, confirming the effectiveness of the inverse design approach.

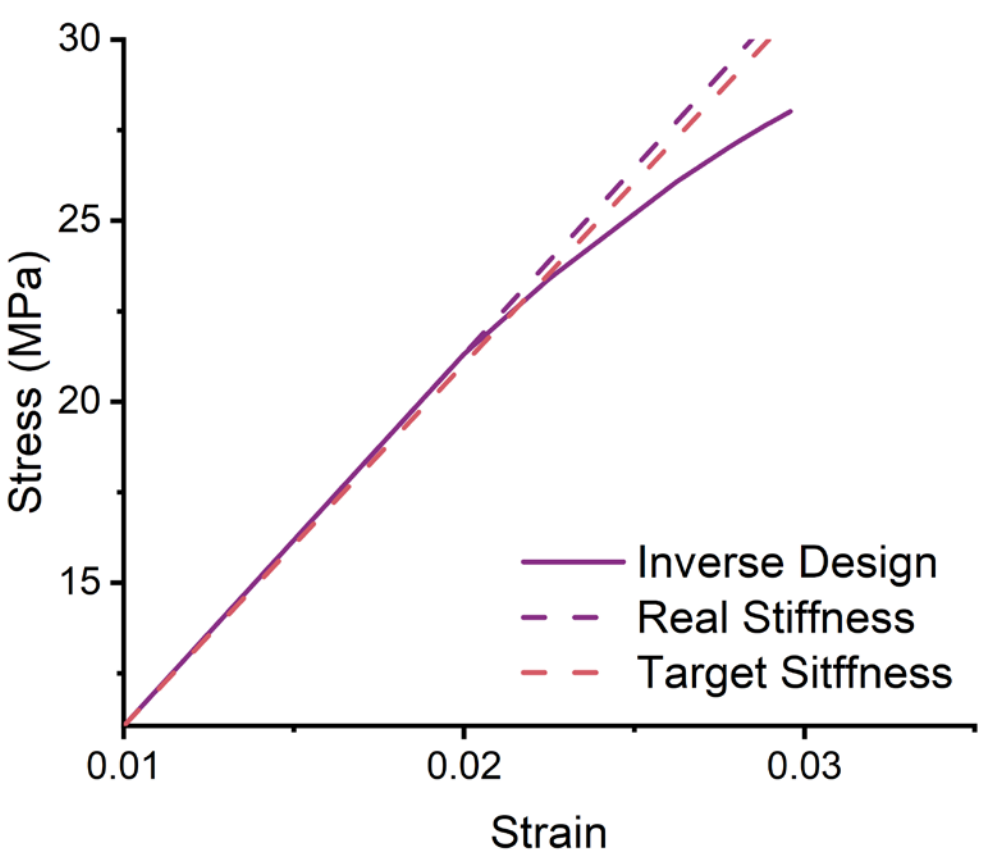

Figure 6. Stress-strain curve of inverse designed structure with target stiffness value of 1000 Mpa.

### 3.4. Extension to nonlinear curved-member lattices and target-shape inverse design

The straight-beam dataset demonstrates topology generation and property learning, but a binary edge representation cannot fully access the large-deformation behaviors needed for programmable shape morphing. Curved beams provide an additional geometric mechanism: under loading, they can rotate, uncoil, bend, and redistribute strain before the base material reaches large local extension. Horseshoe and serpentine microstructures are well established as routes to compliant, highly deformable networks, while more recent work has shown the potential of curved-element metamaterials for nonlinear mechanical programming and inverse design [29-35].

In our current extension, each active graph edge is replaced by a parameterized horseshoe-shaped centerline. Instead of storing only its endpoint pair, each edge carries a vector of geometric parameters - such as span, rise, thickness, and curvature-control variables - together with parameters describing the nonlinear material model. The graph therefore becomes an attributed geometric graph. Connectivity controls how forces are transmitted through the network, while continuous edge attributes tune local compliance and the sequence of geometric nonlinearities.

This representation preserves the main advantage of the growth framework: topology can still be created by the same temporal rules. The only difference is that a newly born edge now has a shape and material state in addition to its endpoints. Consequently, the design space expands smoothly from topology-only straight trusses to coupled topology-geometry nonlinear networks.

The design objective can also be generalized from a single scalar property to a spatially resolved deformation target. Let u*(s) denote a prescribed boundary displacement or target profile under a specified load. A nonlinear analysis maps the edge parameters q to the deformed lattice u(q). The inverse problem then minimizes the mismatch between the deformed boundary and the target while enforcing geometric, stress, manufacturability, or regularity constraints. A generic objective is

$$J(\mathbf{q}) = \sum_k \left\| \mathbf{u}_k(\mathbf{q}) - \mathbf{u}_k^* \right\|^2 + \lambda R(\mathbf{q})$$

where R(q) collects penalty terms and constraints. Depending on the size of the problem, the forward map can be supplied either by nonlinear FEA or by an edge-aware surrogate trained on curved-member simulations. In this form, the same design philosophy supports objectives that cannot be summarized by stiffness alone: a lattice may be required to form an arch, a traveling wave, an asymmetric contour, or a prescribed sequence of intermediate shapes under loading.

Figure 7 summarizes this extension. The validated straight-beam results establish the graph-learning and inverse-screening foundation; the curved-member implementation shows how the framework can be expanded to nonlinear shape programming without abandoning its graph-based representation.

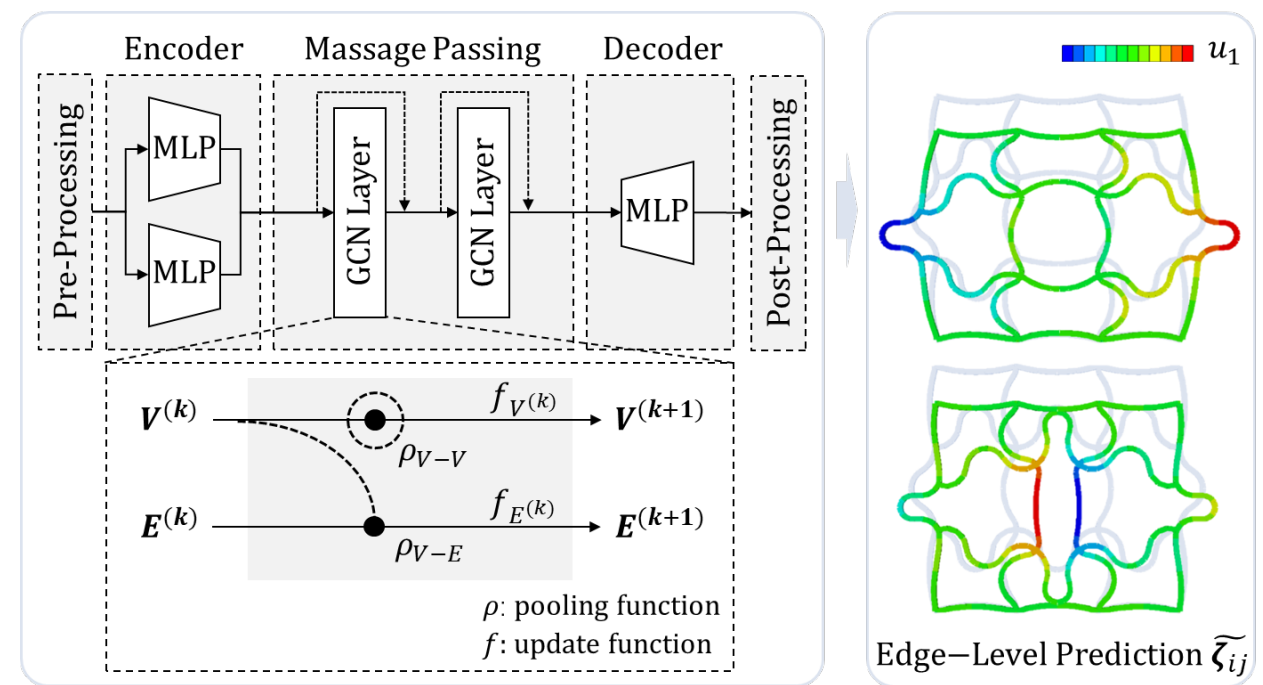

Figure 7: Inverse design of curved lattices.

## 4. Discussion and conclusion

This study develops a growth-inspired representation for mechanical lattice metamaterials that treats structural generation as a temporally ordered morphogenetic process. A discrete dot matrix defines candidate nodes; cross-layer branching creates a hierarchical skeleton; and later intra-layer connections add redundancy and alternative load paths. The same developmental logic can be visualized in a two-dimensional leaf-like domain and transferred directly to a three-dimensional 27-node lattice scaffold.

Using 1,000 generated three-dimensional structures, finite element analysis established a stiffness database with mean 1098 MPa and standard deviation 126.63 MPa. A GCNN learned the resulting topology-property map with $R^2$ = 0.84 and MAE = 37 MPa. In a stiffness-targeting inverse demonstration, a 1000 MPa target produced a design validated at 1027.49 MPa. These results show that a compact growth grammar can generate a learnable yet mechanically diverse structural population and that graph surrogates can substantially reduce the cost of inverse search.

The broader implication is that “growth” can serve as a design coordinate. Once edges are allowed to carry continuous geometry and nonlinear material attributes, the same graph framework naturally extends from straight-beam stiffness design to parameterized horseshoe lattices and prescribed deformation-shape programming. This progression - from developmental generation, to graph learning, to nonlinear shape inverse design - provides a scalable path toward architected materials that are not merely optimized as static geometries but designed as evolving mechanical systems.

## Acknowledgements
The authors acknowledge the research environments and computational support available through their respective institutions.

## Declaration of competing interest
The authors declare that they have no known competing financial interests or personal relationships that could have appeared to influence the work reported in this paper.

## Data availability
Data supporting the findings of this study, including generated lattice structures and model outputs, are available from the corresponding author upon reasonable request.